\documentclass[journal]{IEEEtran}
\usepackage{graphicx,times,amsmath,amssymb,subfigure,stfloats,threeparttable,booktabs,bm,multirow,footnote,cite}
\usepackage[lined,ruled,commentsnumbered]{algorithm2e}
 \allowdisplaybreaks \interdisplaylinepenalty=2500

\makeatletter
\newcommand{\tabincell}[2]{\begin{tabular}{@{}#1@{}}#2\end{tabular}}
\makeatother

\begin{document}
\title{Unsupervised Pool-Based\\ Active Learning for Linear Regression}

\author{Ziang~Liu and Dongrui~Wu
\thanks{Z.~Liu and D.~Wu are with the Key Laboratory of the Ministry of Education for Image Processing and Intelligent Control, School of Artificial Intelligence and Automation, Huazhong University of Science and Technology, Wuhan 430074, China. Email: ziangliu@hust.edu.cn, drwu@hust.edu.cn.}
\thanks{D.~Wu is the corresponding author.}}

\maketitle

\begin{abstract}
In many real-world machine learning applications, unlabeled data can be easily obtained, but it is very time-consuming and/or expensive to label them. So, it is desirable to be able to select the optimal samples to label, so that a good machine learning model can be trained from a minimum amount of labeled data. Active learning (AL) has been widely used for this purpose. However, most existing AL approaches are supervised: they train an initial model from a small amount of labeled samples, query new samples based on the model, and then update the model iteratively. Few of them have considered the completely unsupervised AL problem, i.e., starting from zero, how to optimally select the very first few samples to label, without knowing any label information at all. This problem is very challenging, as no label information can be utilized. This paper studies unsupervised pool-based AL for linear regression problems. We propose a novel AL approach that considers simultaneously the informativeness, representativeness, and diversity, three essential criteria in AL. Extensive experiments on 14 datasets from various application domains, using three different linear regression models (ridge regression, LASSO, and linear support vector regression), demonstrated the effectiveness of our proposed approach.
\end{abstract}


\maketitle

\section{Introduction}

\IEEEPARstart{A} labeled training dataset is often needed in machine learning. However, in many real-world applications, it is relatively easy to obtain unlabeled data, but time-consuming and/or expensive to label them. For example, in emotion estimation from speech signals, it is easy to record a large number of speech utterances, but to evaluate the 3D emotion primitives \cite{Mehrabian1980} (valence, arousal, and dominance) in an utterance, an assessor must listen carefully to it, maybe even multiple times. Moreover, since emotion evaluations are subjective and subtle, usually multiple assessors (e.g., 6-17 in the VAM corpus \cite{Grimm2008}, and at least 110 in IADS-2 \cite{Bradley2007}) are needed for each utterance, which is very time-consuming and labor-intensive. Another example is 180-day post-fracturing cumulative oil production prediction in enhanced oil recovery in the oil and gas industry \cite{drwuSPE2009}. The inputs (fracturing parameters of an oil well, such as its location, length of perforations, number of zones/holes, volumes of injected slurry/water/sand, etc.) can be easily recorded during the fracturing operation, but to get the groundtruth output (180-day post-fracturing cumulative oil production), one has to wait for at least 180 days.

In such applications, one has to determine carefully which unlabeled samples should be selected for labeling.
Active learning (AL) \cite{Settles2009} is a promising solution. In contrast to random selection, it reduces the data labeling effort by querying the samples that are the most useful in improving the model training performance.

AL can be used for both classification and regression. Numerous AL approaches have been proposed for classification in the literature \cite{Settles2009}, but relatively fewer for regression \cite{Burbidge2007,Cai2017,Cai2013,Cohn1996,Freund1997,MacKay1992,Sugiyama2006,Sugiyama2009,Willett2006,drwuEBMAL2016, Yu2010,drwuSAL2019,drwuiGS2019}. These active learning for regression (ALR) approaches are either \emph{population-based} or \emph{pool-based} \cite{Sugiyama2009}. This paper considers the latter, where a pool of unlabeled samples is given, and ALR needs to optimally select some to label, so that a good linear regression model can be trained from them.

Most existing pool-based ALR approaches \cite{Burbidge2007,Cai2013,drwuEBMAL2016,Yu2010,drwuSAL2019,drwuiGS2019} focused on the simpler case that initially there are a small number of labeled samples, so that one can build a regression model from them, and then select more samples for labeling based on the model. Only four studies \cite{Burbidge2007,drwuEBMAL2016,drwuSAL2019,drwuiGS2019} explicitly considered completely unsupervised pool-based ALR (the details of these approaches will be introduced in the next section), i.e., how to optimally select the first a few samples to label, which is also the focus of this paper.

More specifically, we consider the following problem: Given a pool of $N$ unlabeled samples, how to optimally select $M$ from them to label, so that an optimal linear regression model can be built from them? Here $M$ is a small number specified by the user (as $M$ gets larger, the benefit of ALR usually vanishes gradually), and we focus on linear regression only.

We propose a novel informativeness-representativeness-diversity (IRD) ALR approach in this paper, which identifies $M$ initial samples to query, by considering simultaneously the informativeness, representativeness, and diversity, three essential criteria for ALR \cite{drwuSAL2019}. Extensive experiments on 14 datasets from various application domains, using three different linear regression models, demonstrated that the $M$ samples selected by our IRD approach can achieve significantly better performance than those by three state-of-the-art ALR approaches.

The main contributions of this paper are:
\begin{enumerate}
\item We propose the first unsupervised ALR approach which considers simultaneously the informativeness, the representativeness, and the diversity, of the $M$ selected samples, when $M\le d+1$, where $d$ is the feature dimensionality.

\item We propose an iterative approach to select the remaining $M-d-1$ samples, when $M>d+1$, by considering both the representativeness and the diversity.

\item We demonstrate the superior performance of the proposed IRD approach, on various real-world datasets, using three popular linear regression models.
\end{enumerate}

The remainder of this paper is organized as follows: Section~\ref{sect:existing} introduces three existing unsupervised ALR approaches, and points out their limitations. Section~\ref{sect:IRD} proposes our unsupervised pool-based IRD approach. Section~\ref{sect:experiments} describes the 14 datasets to evaluate the effectiveness of IRD, and the corresponding experimental results. Finally, Section~\ref{sect:conclusions} draws conclusions.

\section{Existing Unsupervised Pool-Based ALR Approaches} \label{sect:existing}

Wu \cite{drwuSAL2019} proposed the following three criteria that should be considered in pool-based sequential ALR, which also apply to unsupervised ALR:
\begin{enumerate}
\item \emph{Informativeness}, which could be measured by uncertainty (entropy, distance to the decision boundary, confidence of the prediction, etc.), expected model change, expected error reduction, and so on.

\item \emph{Representativeness}, which could be measured by the number of samples that are similar or close to a target sample. This criterion prevents an outlier from being selected.

\item \emph{Diversity}, which means that the selected samples should scatter across the full input space, so that a good global model can be learned.
\end{enumerate}

Next, we introduce three existing unsupervised pool-based ALR approaches in the literature, and check them against the above three criteria. We assume the pool consists of $N$ $d$-dimensional unlabeled samples $\mathbf{x}_n=[x_{n,1},x_{n,2},\ldots,x_{n,d}]^T\in\mathbb{R}^{d\times 1}$, $n=1,2,...,N$, and the user wants to select $M$ of them to label.

\subsection{P-ALICE} \label{sect:P-ALICE}

Sugiyama and Nakajima \cite{Sugiyama2009} proposed Pool-based Active Learning using the Importance-weighted least-squares learning based on Conditional Expectation of the generalization error (P-ALICE), a completely unsupervised ALR approach to select the initial few samples to label. Its main idea is to identify $M$ samples and their associated weights, so that a weighted linear regression model constructed from them can minimize the estimated mean squared loss on the $N$ samples, by considering the covariate shift between the training and test samples.

Let
\begin{align}
\mathbf{U}=\frac{1}{N}\sum_{n=1}^N \mathbf{x}_n\mathbf{x}_n^T\in \mathbb{R}^{d\times d},
\end{align}
$\mathbf{U}^{-1}\in \mathbb{R}^{d\times d}$ be the inverse of $\mathbf{U}$, and $\mathbf{U}^{-1}_{i,j}$ be the $(i,j)$th element of $\mathbf{U}^{-1}$. P-ALICE first defines a family of resampling bias functions parameterized by a scalar $\lambda\in[0,1]$:
\begin{align}
b^\lambda(\mathbf{x}_n)=\left(\sum_{i,j=1}^d \mathbf{U}^{-1}_{i,j}x_{n,i}x_{n,j}\right)^\lambda
\end{align}
For each different $\lambda$, it selects $M$ unlabeled samples from the pool with probability proportional to $b^\lambda(\mathbf{x}_n)$. Denote the selected samples as $\{\mathbf{x}_m^\lambda\}_{m=1}^M$. Then, the mean squared loss on the $N$ samples is estimated as
\begin{align}
Q(\lambda)=trace[\mathbf{UL}^\lambda(\mathbf{L}^\lambda)^T],
\end{align}
where
\begin{align}
\mathbf{L}&=[\mathbf{X}^\lambda\mathbf{W}^\lambda(\mathbf{X}^\lambda)^T]^{-1} \mathbf{X}^\lambda\mathbf{W}^\lambda \in \mathbb{R}^{d\times M},
\end{align}
in which
\begin{align}
\mathbf{X}^\lambda &=[\mathbf{x}_1^\lambda,\mathbf{x}_2^\lambda,\ldots,\mathbf{x}_M^\lambda] \in\mathbb{R}^{d\times M}\\
\mathbf{W}^\lambda &=diag([b^\lambda(\mathbf{x}_1^\lambda),b^\lambda(\mathbf{x}_2^\lambda),\ldots, b^\lambda(\mathbf{x}_M^\lambda)]^{-1}) \in \mathbb{R}^{M\times M}
\end{align}
P-ALICE then identifies $\lambda^*=\arg\min_{\lambda} Q(\lambda)$, i.e., $\lambda$ that results in the smallest mean squared loss on the $N$ samples, and selects the corresponding $\{\mathbf{x}^{\lambda^*}_m\}_{m=1}^M$ for labeling. Because each such $\mathbf{x}^{\lambda^*}_i$ is associated with a weight $b^{\lambda^*}(\mathbf{x}^{\lambda^*}_i)$, P-ALICE finally computes a weighted linear regression model to accommodate the covariate shift between the training and the test samples.

Checking against the three criteria for ALR, P-ALICE explicitly considers the informativeness (the estimated mean squared loss), but not the representativeness and the diversity.

\subsection{Greedy Sampling in the Input Space (GSx)}\label{sect:GSx}

Yu and Kim \cite{Yu2010} proposed greedy sampling (GS), a completely unsupervised pool-based ALR approach. Given a non-empty seed set, GS can select additional unlabeled samples without using any label information. However, GS needs to have at least one unlabeled sample as the seed first, and \cite{Yu2010} did not explicitly explain how the first seed should be identified. Wu \emph{et al.} \cite{drwuiGS2019} proposed GSx, which slightly improves GS by specifying the first sample as the one closest to the centroid of all $N$ unlabeled samples. More details on GSx are introduced next.

Without loss of generality, assume the first $M_0$ samples have been selected by GSx. For each of the $N-M_0$ remaining unlabeled samples ${\{\mathbf{x}_n\}}^N_{n=M_0+1}$, GSx computes its distance to each of the $M_0$ selected samples, i.e.,
\begin{align}\label{E_GSx}
d_{nm}&=||\bm{x}_{n}-\bm{x}_{m}||,\quad m=1,...,M_0;\ n=M_0+1,...,N
\end{align}
Then, it computes $\underline{d}_n$ as the minimum distance from $\bm{x}_{n}$ to the $M_0$ selected samples, i.e.,
\begin{align}
\underline{d}_n=\min_{m}d_{nm},\quad n=M_0+1,...,N
\end{align}
and selects the sample with the maximum $\underline{d}_n$ to label. This process repeats until all $M$ samples have been selected.

Checking against the three criteria for ALR, GSx only considers the diversity, but not the informativeness and the representativeness at all.

\subsection{Representativeness-Diversity (RD)} \label{sect:RD}

Wu \cite{drwuSAL2019} proposed a new pool-based ALR approach, by considering particularly the representativeness and the diversity of the selected samples, so it is denoted as RD.

RD has two parts, one for initialization (unsupervised), and the other for subsequent iterations after the initially selected samples are labeled (supervised). The unsupervised part of RD first performs $k$-means ($k=d+1$) clustering on the $N$ unlabeled samples, and then selects from each cluster the sample closest to its centroid for labeling. This idea had also been used in \cite{drwuEBMAL2016}.

As its name suggests, RD only considers the representativeness and the diversity in its initialization, but not the informativeness at all.

\subsection{Summary}

Table~\ref{tab:IRD} summarizes the main characteristics of P-ALICE, GSx, and RD. Each of them only explicitly considers a subset of the three essential criteria for ALR; so, there is still room for improvement.

\begin{table}[h] \centering \setlength{\tabcolsep}{1mm}
\caption{Criteria considered in the three existing and the proposed unsupervised pool-based ALR approaches.}   \label{tab:IRD}
\begin{tabular}{c|c|ccc}   \hline
&Approach & Informativeness & Representativeness & Diversity \\ \hline
&P-ALICE & $\checkmark$ & $-$&$-$ \\
Existing & GSx & $-$& $-$& $\checkmark$\\
& RD & $-$&$\checkmark$ & $\checkmark$ \\ \hline
Proposed & IRD &$\checkmark$ & $\checkmark$ &$\checkmark$ \\ \hline
\end{tabular}
\end{table}

\section{IRD} \label{sect:IRD}

This section introduces our proposed IRD approach for unsupervised pool-based ALR. As its name suggests, IRD considers informativeness, representativeness, and diversity simultaneously.

Let $M$ be the number of samples to be selected, and $d$ the feature dimensionality. We consider three cases in IRD separately: $M=d+1$, $M<d+1$, and $M>d+1$.

\subsection{Case 1: $M=d+1$}

\begin{figure}[htpb]\centering
\includegraphics[width=\linewidth,clip]{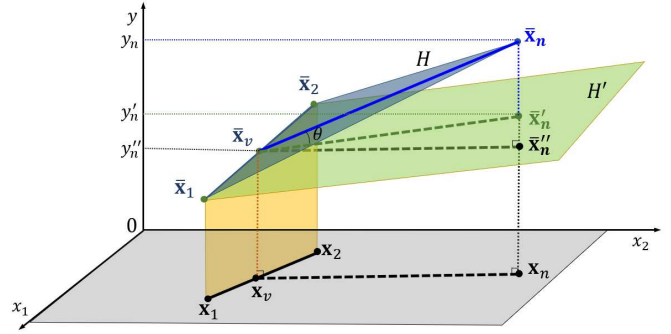}
\caption{Illustration of IRD when $d=2$.} \label{fig:IRD}
\end{figure}

For $d$ features, generally we need to select at least $d+1$ samples in order to construct a linear regression model $f(\mathbf{x})=\mathbf{x}^T\mathbf{w}+b$, where $\mathbf{w}\in \mathbb{R}^{d\times 1}$ consists of the regression coefficients, and $b$ is the bias. We will start with a specific example with $d=2$ to illustrate the basic idea of IRD (Fig.~\ref{fig:IRD}).

Assume the first two unlabeled samples, $\mathbf{x}_1$ and $\mathbf{x}_2$, have been selected, and now we need to select the third sample from $\{\mathbf{x}_n\}_{n=3}^N$. For the convenience of illustration, we denote $\bar{\mathbf{x}}_n=[\mathbf{x}_n; y_n]\in\mathbb{R}^{(d+1)\times 1}$, $n=1,...,N$.

Let $H'$ be the $d$-dimension optimal manifold that passes through $\bar{\mathbf{x}}_1$ and $\bar{\mathbf{x}}_2$, and best fits the remaining $N-2$ samples. In this completely unsupervised problem, we do not know $H'$; however, if we are given all $N$ $\bar{\mathbf{x}}_n$, and the requirement that $H'$ must pass through $\bar{\mathbf{x}}_1$ and $\bar{\mathbf{x}}_2$, there always exists such an $H'$.

Any particular $\bar{\mathbf{x}}_n$ ($n=3,...,N$) and $\{\bar{\mathbf{x}}_1,\bar{\mathbf{x}}_2\}$ can determine a 2-dimension manifold $H$, as shown in Fig.~\ref{fig:IRD}. $H$ and $H'$ intersects at the line $\overrightarrow{\bar{\mathbf{x}}_1\bar{\mathbf{x}}_2}$. Also, for any particular $\bar{\mathbf{x}}_n$ ($n=3,...,N$), we can find a point $\bar{\mathbf{x}}_v$ on the line $\overrightarrow{\bar{\mathbf{x}}_1\bar{\mathbf{x}}_2}$ such that $\overrightarrow{\bar{\mathbf{x}}_1\bar{\mathbf{x}}_2}\perp \overrightarrow{\bar{\mathbf{x}}_v\bar{\mathbf{x}}_n}$, where $\bar{\mathbf{x}}_v\bar{\mathbf{x}}_n\in H'$, as shown in Fig.~\ref{fig:IRD}.

For optimal performance, we want $H$ to be as close to $H'$ as possible. Next we show how to identify the optimal $\mathbf{x}_n^*$ for this purpose.

Let $\overrightarrow{\bar{\mathbf{x}}_n\bar{\mathbf{x}}_n'}$ be the line that passes through $\bar{\mathbf{x}}_n$, is parallel to the $y$ axis, and intersects $H'$ at $\bar{\mathbf{x}}_n'$. Then, we can use the angle $\theta$ between $\overrightarrow{\bar{\mathbf{x}}_v\bar{\mathbf{x}}_n}$ and $\overrightarrow{\bar{\mathbf{x}}_v\bar{\mathbf{x}}_n'}$ to represent the closeness\footnote{Note that $\theta$ is not the true angle between $H$ and $H'$, which requires $\overrightarrow{\bar{\mathbf{x}}_v\bar{\mathbf{x}}_n'}\perp\overrightarrow{\bar{\mathbf{x}}_1\bar{\mathbf{x}}_2}$, which generally does not hold for $\bar{\mathbf{x}}_n'$. However, just as the true angle between $H$ and $H'$, $\theta$ decreases monotonically as $H'$ and $H$ get closer to each other, so it can be used to measure the closeness between them. We use such a $\theta$ instead of the true angle between $H$ and $H'$ because this $\theta$ makes our derivation much easier.} of $H$ to $H'$.

From simple geometry in Fig.~\ref{fig:IRD}, we have:
\begin{align}
|\theta|\propto \frac{|\bar{\mathbf{x}}_n-\bar{\mathbf{x}}_n'|}{|\bar{\mathbf{x}}_v-\bar{\mathbf{x}}_n'|}
=\frac{|y_n-y_n'|}{\sqrt{(y_n'-y_n'')^2+|\bar{\mathbf{x}}_v-\bar{\mathbf{x}}_n''|^2}},
\end{align}

The numerator $|y_n-y_n'|$ depends on $y_n$ and $y_n'$, which are completely unknown in our problem setting; so, we have to ignore it. The denominator has two terms. The first term $y_n'-y_n''$ is again completely unknown to us, so we also have to ignore it. The second term $|\bar{\mathbf{x}}_v-\bar{\mathbf{x}}_n''|$ can be computed in unsupervised ALR. Because $\bar{\mathbf{x}}_v$ and $\bar{\mathbf{x}}_n''$ have the same $y$, the distance $|\bar{\mathbf{x}}_v-\bar{\mathbf{x}}_n''|$ is irrelevant to $y$, and equals $|\mathbf{x}_v-\mathbf{x}_n|$, the distance from $\mathbf{x}_n$ to the line $\overrightarrow{\mathbf{x}_1\mathbf{x}_2}$, as illustrated in Fig.~\ref{fig:IRD}.

So, based on the above derivation and all information one can use in completely unsupervised pool-based ALR, we \emph{approximately} have:
\begin{align}
|\theta|\propto \frac{1}{|\mathbf{x}_v-\mathbf{x}_n|} \label{eq:ID}
\end{align}
Note that (\ref{eq:ID}) is derived from the motivation that $H$ and $H'$ need to be as close as possible, so it considers the informativeness of $\mathbf{x}_n$. Additionally, $|\mathbf{x}_v-\mathbf{x}_n|$ can also be viewed as a distance from $\mathbf{x}_n$ to those samples that have already be selected ($\bar{\mathbf{x}}_1$ and $\bar{\mathbf{x}}_2$ in our example). To make $\theta$ small, we need to make sure $|\mathbf{x}_v-\mathbf{x}_n|$ is large, i.e., (\ref{eq:ID}) also ensures the diversity among the selected samples. In summary, (\ref{eq:ID}) considers simultaneously the informativeness and the diversity in selecting the third sample.

However, if only (\ref{eq:ID}) is used in selecting the third sample, it will always select the sample that is farthest away from the line $\overrightarrow{\mathbf{x}_1\mathbf{x}_2}$, which could be an outlier. To consider also the representativeness, we compute the mean distance from $\mathbf{x}_n$ to all $N$ samples, and select the optimal sample to label as\footnote{We have also considered other approaches to combine the representativeness with the informativeness and the diversity, e.g., $\mathbf{x}_n^*=\arg\min_{\mathbf{x}_n}(\frac{1}{N}\sum_{i=1}^N|\mathbf{x}_i-\mathbf{x}_n|^2)^{1/2}+\lambda\cdot
|\mathbf{x}_v-\mathbf{x}_n|$; however, this introduces an extra hyper-parameter $\lambda$, and our experiments showed that performance obtained from the best $\lambda$ was worse than that from (\ref{eq:IRD2}). So, we eventually used (\ref{eq:IRD2}) for its simplicity and accuracy.}:
\begin{align}
\mathbf{x}_n^*=\arg\min_{\mathbf{x}_n}
\frac{\left(\frac{1}{N}\sum_{i=1}^N\left|\mathbf{x}_i-\mathbf{x}_n\right|^2\right)^{\frac{1}{2}}}
{|\mathbf{x}_v-\mathbf{x}_n|} \label{eq:IRD2}
\end{align}

When $d>2$, we can have similar derivations, by replacing the line $\overrightarrow{\mathbf{x}_1\mathbf{x}_2}$ with the $(d-1)$-dimension manifold $C$, which passes through all $d$ already selected samples, $\{\bar{\mathbf{x}}_t\}_{t=1}^d$. Then, (\ref{eq:IRD2}) becomes
\begin{align}
\mathbf{x}_n^*=\arg\min_{\mathbf{x}_n}
\frac{\left(\frac{1}{N}\sum_{i=1}^N|\mathbf{x}_i-\mathbf{x}_n|^2\right)^{\frac{1}{2}}}
{dist(\mathbf{x}_n,C)} \label{eq:IRD}
\end{align}
where $dist(\mathbf{x}_n,C)$ is the distance from $\mathbf{x}_n$ to the manifold $C$.

To compute $dist(\mathbf{x}_n,C)$, we need to first find a vector $\mathbf{w}\in \mathbb{R}^{d\times 1}$ perpendicular to $C$, i.e.,
\begin{align}
\left[\begin{array}{cccc}
        \mathbf{x}_1 & \mathbf{x}_2 & \ldots & \mathbf{x}_d \\
        1 & 1 & \ldots & 1
      \end{array}\right]^T
      \left[\begin{array}{c}\mathbf{w}\\
      b
      \end{array}\right]=\mathbf{0}.
\end{align}
Then,
\begin{align}
dist(\mathbf{x}_n,C)=\frac{\mathbf{x}_n^T\mathbf{w}+b}{|\mathbf{w}|}. \label{eq:dist}
\end{align}

To our knowledge, formulas like (\ref{eq:IRD}) have never appeared in ALR. In AL for classification, there are approaches \cite{Tong2001,Lewis1994b,Lewis1994} that select the samples closest to the current classification boundary (i.e., samples with the maximum uncertainty) for labeling, but they are supervised: some labeled samples must be available first to initialize the classifier. Our approach (\ref{eq:IRD}) is completely unsupervised, and it is for linear regression instead of classification.

\subsection{Case 1: Iterative Improvement}

The above approach selects the $(d+1)$th sample, given that the first $d$ samples have been selected. The optimality of the $(d+1)$th sample also depends on the optimality of the first $d$ samples.

Next we propose an expectation-maximization (EM) approach to optimize the $d+1$ samples iteratively: we first select the $d$ samples by GSx or RD, fix them, and then select the $(d+1)$th sample by using (\ref{eq:IRD}). Then, we use (\ref{eq:IRD}) repeatedly to optimize each sample $\mathbf{x}_t$ ($t=1,...,d+1$), by fixing the remaining $d$ samples. This process iterates until the selected samples converge, or the maximum number of iterations is reached.

In summary, the pseudo-code for the case $M=d+1$ is given in Algorithm~\ref{alg:1}.

\begin{algorithm}[htbp] 
\KwIn{$N$ unlabeled samples, $\{\mathbf{x}_n\}^N_{n=1}$, where $\mathbf{x}_n\in\mathbb{R}^{d\times 1}$\;
    \hspace*{9mm} $c_{\max}$, the maximum number of iterations.}
\KwOut{$\{\mathbf{x}_t\}_{t=1}^M$, the optimal set of $M=d+1$ samples to label.}
Use GSx or RD to initialize the $M$ samples\;
Save the indices of the $M$ samples to the first row of matrix $P$\;
$c=0$\;
\While{$c<c_{\max}$}{
    Denote the $M$ selected samples as $\{\mathbf{x}_t\}_{t=1}^M$, and the remaining samples as $\{\mathbf{x}_n\}_{n=M+1}^N$\;
    \For{$t=1,...,M$}{
        Fix $\{\mathbf{x}_1,\ldots,\mathbf{x}_{t-1},\mathbf{x}_{t+1},\ldots,\mathbf{x}_M\}$ as the $d$ points on Manifold $C$\;
        Compute $dist(\mathbf{x}_n,C)$ in (\ref{eq:dist}) for each $\mathbf{x}_n$, $n=M+1,\ldots,N$\;
        Set $\mathbf{x}_t$ to $\mathbf{x}_n^*$ computed from (\ref{eq:IRD})\;}
    \uIf{the indices of the $M$ samples match any row in $P$}{
        \textbf{Break}\;}
    \Else{
        Save the indices of the $M$ samples to the next row of $P$\;}
            $c=c+1$\;}
\caption{The IRD algorithm for $M=d+1$.} \label{alg:1}
\end{algorithm}

\subsection{Case 2: $M<d+1$}

Case~1 considers the scenario that $M=d+1$, i.e., the number of selected samples equals the number of the features plus one. However, this is a very special case, and in practice $M$ could be smaller than $d+1$. In this case, the $(d-1)$-dimension manifold $C$ in (\ref{eq:IRD}) cannot be uniquely determined, and hence $\mathbf{x}_n^*$ cannot be identified directly by using (\ref{eq:IRD}).

We propose a new approach for handling $M<d+1$. We first perform principal component analysis (PCA) on all $N$ $\mathbf{x}_n$ to identify the $M-1$ leading principal components, and then replace each $\mathbf{x}_n$ by the corresponding $M-1$ scores. (\ref{eq:IRD}) can then be used again on these transformed $\mathbf{x}_n$. 

The pseudo-code for $M<d+1$ is given in Algorithm~\ref{alg:2}.

\begin{algorithm}[htbp] 
\KwIn{$N$ unlabeled samples, $\{\mathbf{x}_n\}^N_{n=1}$, where $\mathbf{x}_n\in\mathbb{R}^{d\times 1}$\;
\hspace*{9mm} $M$, the number of samples to be selected ($M<d+1$)\;
    \hspace*{9mm} $c_{\max}$, the maximum number of iterations.}
\KwOut{$\{\mathbf{x}_t\}_{t=1}^M$, the optimal set of samples to label.}
Perform PCA on $[\mathbf{x}_1\ \mathbf{x}_2\ \ldots \mathbf{x}_N]^T\in R^{N\times d}$\;
Replace each $\mathbf{x}_n\in R^{d\times 1}$ by its $M-1$ leading scores in the PCA\;
Use Algorithm~\ref{alg:1} to identify the $M$ samples.
\caption{The IRD algorithm for $M<d+1$.} \label{alg:2}
\end{algorithm}

\subsection{Case 3: $M>d+1$}

When $M>d+1$, we first initialize $d+1$ samples using the approach proposed for Case~1. Then, we fix these $d+1$ samples, and proceed to identify the remaining $M-d-1$ samples.

We use $k$-means ($k=M-d-1$) clustering to cluster the remaining $N-d-1$ samples into $M-d-1$ clusters, and then select one sample from each cluster, similar to the RD approach. However, here we have a small improvement: instead of selecting the sample closest to the centroid of each cluster, we use an iterative approach to select the remaining $M-d-1$ samples.

Without loss of generality, assume the first $d+1$ samples are those selected by using Algorithm~1, the next $M-d-2$ samples have been temporally fixed (e.g., as the sample closest to its corresponding cluster centroid), and we want to optimize the $M$th sample belonging to the $(M-d-1)$th cluster. For each $\mathbf{x}_n$ in this cluster, we compute its representativeness as the inverse of its mean distance to other samples in the same cluster. Let $S$ be the indices of the samples in the $(M-d-1)$th cluster. Then, the representativeness of $\mathbf{x}_n$ is
\begin{align}
R(\mathbf{x}_n)=\frac{|S|}{\sum_{i\in S}|\mathbf{x}_n-\mathbf{x}_i|^2}
\end{align}
where $|S|$ is the number of elements in $S$.

We compute the diversity of $\mathbf{x}_n$ as its minimum distance to the $M-1$ selected samples, i.e.,
\begin{align}
D(\mathbf{x}_n)=\min_{t=1,...,M-1} |\mathbf{x}_n-\mathbf{x}_t|
\end{align}
And finally, the combined effect of the representativeness and the diversity is computed as:
\begin{align}
RD(\mathbf{x}_n)=R(\mathbf{x}_n)\cdot D(\mathbf{x}_n) \label{eq:IRDr}
\end{align}
We then select
\begin{align}
\mathbf{x}_n^*=\arg\max_{\mathbf{x}_n} RD(\mathbf{x}_n) \label{eq:RD}
\end{align}
to replace the $M$th sample. We repeat this process for each of $\{\mathbf{x}_t\}_{t=d+2}^M$, until none of them would change, or the maximum number of iterations has reached.

\begin{algorithm}[htbp] 
\KwIn{$N$ unlabeled samples, $\{\mathbf{x}_n\}^N_{n=1}$, where $\mathbf{x}_n\in\mathbb{R}^{d\times 1}$\;
\hspace*{9mm} $M$, the number of samples need to be selected ($M>d+1$)\;
    \hspace*{9mm} $c_{\max}$, the maximum number of iterations.}
\KwOut{$\{\mathbf{x}_t\}_{t=1}^M$, the optimal set of $M$ samples to label.}
Use Algorithm~\ref{alg:1} to identify the first $d+1$ samples, and assign them to $\{\mathbf{x}_t\}_{t=1}^{d+1}$\;
Perform $k$-means ($k=M-d-1$) clustering on the remaining $N-d-1$ samples\;
Initialize $\mathbf{x}_t$ as the sample closest to the centroid of the $(t-d-1)$th cluster, $t=d+2,...,M$\;
Save the indices of the $M$ samples to the first row of matrix $P$\;
$c=0$\;
\While{$c<c_{\max}$}{
    \For{$t=d+2,...,M$}{
        Set $\mathbf{x}_t$ to $\mathbf{x}_n^*$ computed from (\ref{eq:RD})\;}
    \uIf{the indices of the $M$ samples match any row in $P$}{
        \textbf{Break}\;}
    \Else{
        Save the indices of the $M$ samples to the next row of $P$\;}
            $c=c+1$\;}
\caption{The IRD algorithm for $M>d+1$.} \label{alg:3}
\end{algorithm}

\section{Experiments and Results}\label{sect:experiments}

Extensive experiments are performed in this section to demonstrate the performance of the proposed unsupervised pool-based IRD ALR approach.

\subsection{Datasets}

Fourteen datasets from various application domains were used in our study. Their summary is given in Table~\ref{tab:Datasets}.

\begin{table}[htbp] \centering  \setlength{\tabcolsep}{0.5mm}
\caption{Summary of the 14 regression datasets.}   \label{tab:Datasets}
\begin{tabular}{c|cccccc}   \hline
Dataset    &Source &\tabincell{c}{No. of\\samples} &\tabincell{c}{No. of\\raw\\features} &\tabincell{c}{No. of\\numerical\\features} &\tabincell{c}{No. of\\categorical\\features} &\tabincell{c}{No. of\\total\\features}    \\ \hline
Concrete-CS$^1$     &UCI     &103            &7                   &7                         &0           &7\\
Yacht$^2$           &UCI     &308            &6                   &6                         &0           &6\\
autoMPG$^3$         &UCI     &392            &7                   &6                         &1           &9\\
NO2$^4$             &StatLib &500            &7                   &7                         &0           &7\\
Housing$^5$         &UCI     &506            &13                  &13                        &0          &13 \\
CPS$^6$             &StatLib &534            &10                  &7                         &3          &19\\
EE-Cooling$^7$      &UCI     &768            &7                   &7                         &0          &7\\
VAM-Arousal$^8$    &ICME      &947            &46                  &46                        &0         &46\\
Concrete$^9$       &UCI       &1030           &8                   &8                         &0         &8\\
Airfoil$^{10}$        &UCI     &1503           &5                   &5                         &0        &5\\
Wine-Red$^{11}$       &UCI     &1599           &11                  &11                        &0        &11\\
Wine-White$^{11}$     &UCI     &4898           &11                  &11                        &0         &11\\ \hline
\end{tabular}\\ \raggedright
\footnotesize{$^1$ https://archive.ics.uci.edu/ml/datasets/Concrete+Slump+Test}\\
\footnotesize{$^2$ https://archive.ics.uci.edu/ml/datasets/Yacht+Hydrodynamics}\\
\footnotesize{$^3$ https://archive.ics.uci.edu/ml/datasets/auto+mpg}\\
\footnotesize{$^4$ http://lib.stat.cmu.edu/datasets/}\\
\footnotesize{$^5$ https://archive.ics.uci.edu/ml/machine-learning-databases/housing/}\\
\footnotesize{$^6$ http://lib.stat.cmu.edu/datasets/CPS\_85\_Wages}\\
\footnotesize{$^7$ http://archive.ics.uci.edu/ml/datasets/energy+efficiency}\\
\footnotesize{$^8$ https://dblp.uni-trier.de/db/conf/icmcs/icme2008.html}\\
\footnotesize{$^9$ https://archive.ics.uci.edu/ml/datasets/Concrete+Compressive+Strength}\\
\footnotesize{$^{10}$ https://archive.ics.uci.edu/ml/datasets/Airfoil+Self-Noise}\\
\footnotesize{$^{11}$ https://archive.ics.uci.edu/ml/datasets/Wine+Quality}
\end{table}

We used nine datasets from the UCI Machine Learning Repository\footnote{http://archive.ics.uci.edu/ml/index.php}, and two from the CMU StatLib Datasets Archive\footnote{http://lib.stat.cmu.edu/datasets/}, which have also been used in previous ALR experiments \cite{Cai2017,Cai2013,Yu2010,drwuSAL2019,drwuiGS2019}. Two datasets (autoMPG and CPS) contain both numerical and categorical features. For them, we used one-hot coding to covert the categorical values into numerical values before ALR, as in \cite{drwuSAL2019}.

We also used a publicly available affective computing datasets: The \emph{Vera am Mittag} (VAM; \emph{Vera at Noon} in English) German Audio-Visual Spontaneous Speech Database \cite{Grimm2008}, which has been used in many previous studies \cite{Grimm2007b,Grimm2007a,drwuICME2010,drwuInterSpeech2010,drwuMTALR2020}. It contains 947 emotional utterances from 47 speakers. 46 acoustic features \cite{drwuICME2010,drwuInterSpeech2010}, including nine pitch features, five duration features, six energy features, and 26 MFCC features, were extracted, to predict three emotion primitives (arousal, valence, and dominance). Only arousal was considered as the regression output in our experiments.

For each dataset, we normalized each dimension of the input to mean zero and standard deviation one.

\subsection{Algorithms}

We compared the performance of IRD ($c_{\max}=5$) with the following four sampling approaches:
\begin{enumerate}
\item Random sampling (RS), which randomly selects $M$ samples for labeling.
\item P-ALICE, which has been introduced in Section~\ref{sect:P-ALICE}. The parameter $\lambda$ was chosen as the best one from $\{0,.1,.2,.3,.4,.41,.42,...,.59,.6,.7,.8,.9,1\}$, as in \cite{Sugiyama2009}.
\item GSx, which has been introduced in Section~\ref{sect:GSx}.
\item RD, which has been introduced in Section~\ref{sect:RD}.
\end{enumerate}

\subsection{Evaluation Process}\label{sect:evaluation}

For each dataset, we randomly picked 50\% samples as our training pool, and the remaining 50\% as the test set. Each approach selected $M\in[5,15]$ samples from the completely unlabeled training pool for labeling, and then built a linear regression model from them.
The model was then evaluated on the test set, using root mean squared error (RMSE) and correlation coefficient (CC) as the performance measures. This process was repeated 100 times on each dataset to get statistically meaningful results.

Three different linear regression models were trained from the selected samples from each sample selection approach\footnote{We also investigated the ordinary least squares (OLS) linear regression model, which does not have parameter regularization. IRD also achieved the best performance among the five approaches. However, since OLS is very unstable when the number of training samples is small, and hence it is not a rational choice in practice, we do not present its results in this paper.}:

\begin{enumerate}
\item Ridge regression (RR), with the L2 regularization coefficient $\lambda=0.5$. We used a large $\lambda$ to reduce the variance of the regression model, since the number of selected samples was small.
\item LASSO, with the L1 regularization coefficient $\lambda=0.5$.
\item Linear support vector regression (SVR), with $\epsilon=0.1\cdot std(y)$ (half the width of epsilon-insensitive band, where $std(y)$ means the standard deviation of the true label of the $M$ selected sample) and the box constraint $C=1$. SVR comes with L2 regularization term, and the equivalent regularization coefficient is equal to $\frac{1}{2C}$, which has the same magnitude as those in RR and LASSO.
\end{enumerate}

We mainly report results from the RR model in the following subsections, because its RMSEs and CCs were generally better and more stable than those from LASSO and linear SVR, especially for the RS approach. However, as shown in Section~\ref{sect:LASSOSVR}, the relative improvement of IRD over other sampling approaches, especially RS, may be even larger when LASSO or linear SVR was used.

\subsection{Experimental Results Using RR} \label{sect:results_Ridge}

\begin{figure*}[htbp]\centering
\includegraphics[width=\linewidth,clip]{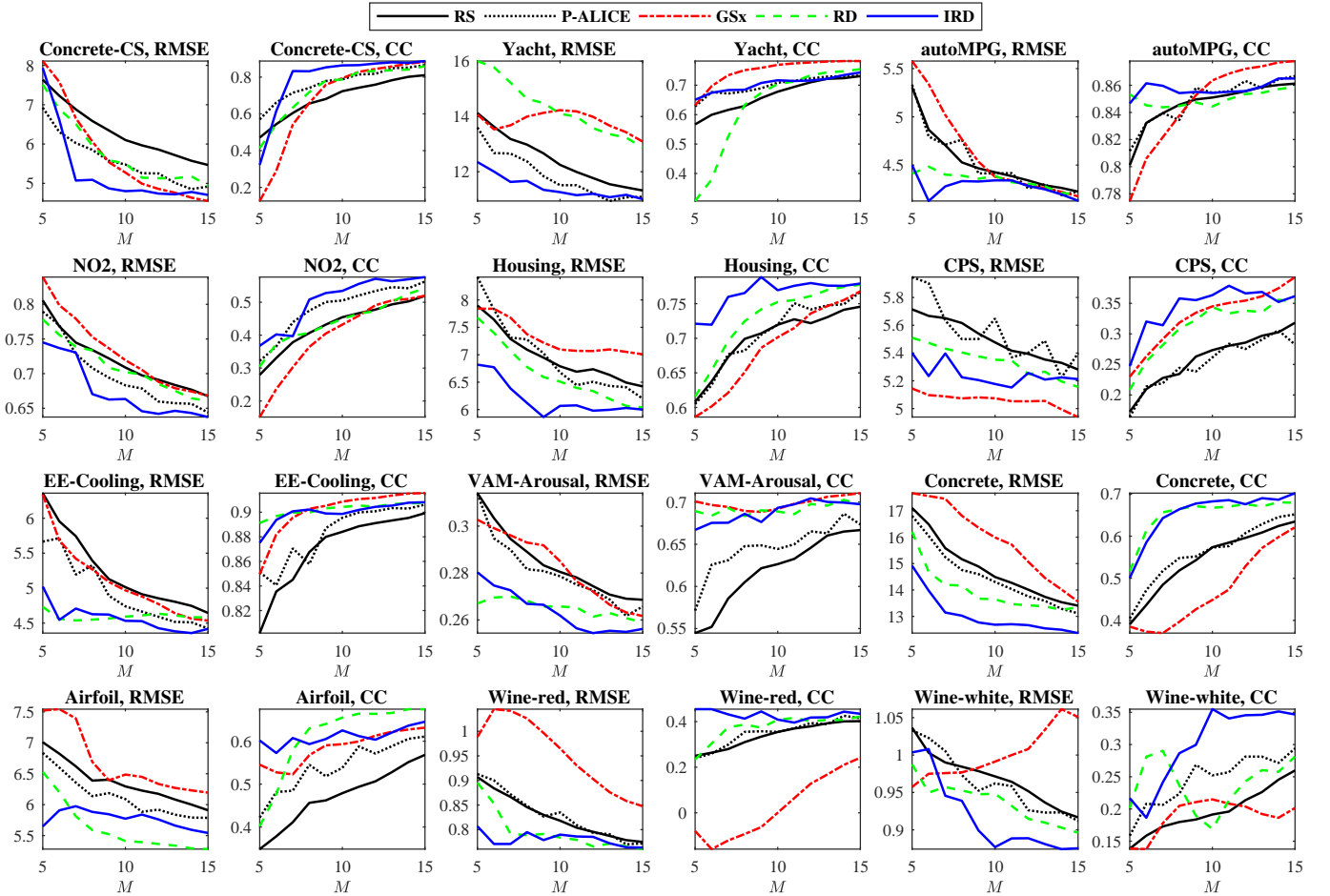}
\caption{Mean of the RMSEs and the CCs on the 12 datasets, averaged over 100 runs. RR ($\lambda=0.5$) was used as the regression model.}\label{fig:data12}
\end{figure*}

The mean RMSEs and CCs of the five sampling approaches on the 12 datasets are shown in Fig.~\ref{fig:data12}, when RR was used as the regression model\footnote{Due to the page limit, we only show the detailed results from RR here, as it generally performed the best among the three regression models. The results from the other two regression models are similar.}.

Generally, as $M$ increased, the RMSEs and CCs of all five sampling approaches improved, which is reasonable, as more labeled training samples generally result in better regression performance. However, there may still be some fluctuations, especially when the number of samples was small. This is because there is a lot of randomness and uncertainty in training a linear regression model from only a few labeled samples.

On most datasets and for most $M$, RS and GSx gave larger RMSEs and smaller CCs than other three approaches, i.e., they had worse performance than the other three approaches. On the contrary, IRD achieved the smallest RMSE and the largest CC on most datasets and for most $M$, indicating that it was the best-performing sample selection approach among the five.

\begin{figure}[htbp]\centering
\subfigure[]{\label{fig:barRR}   \includegraphics[width=\linewidth,clip]{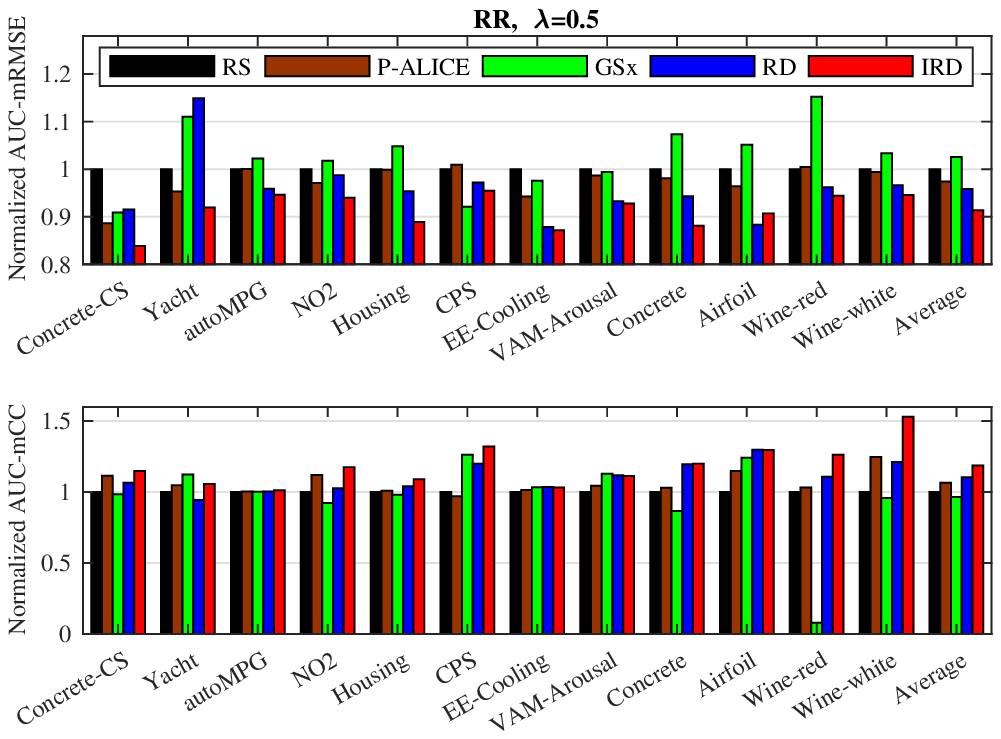}}
\subfigure[]{\label{fig:barLASSO}   \includegraphics[width=\linewidth,clip]{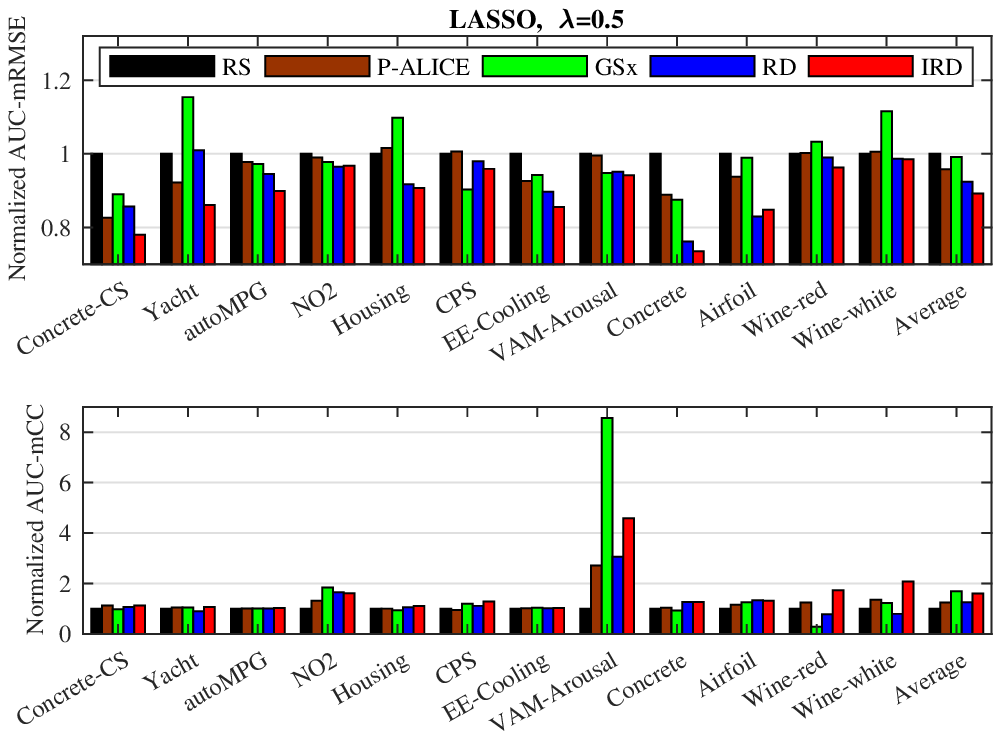}}
\subfigure[]{\label{fig:barSVR}   \includegraphics[width=\linewidth,clip]{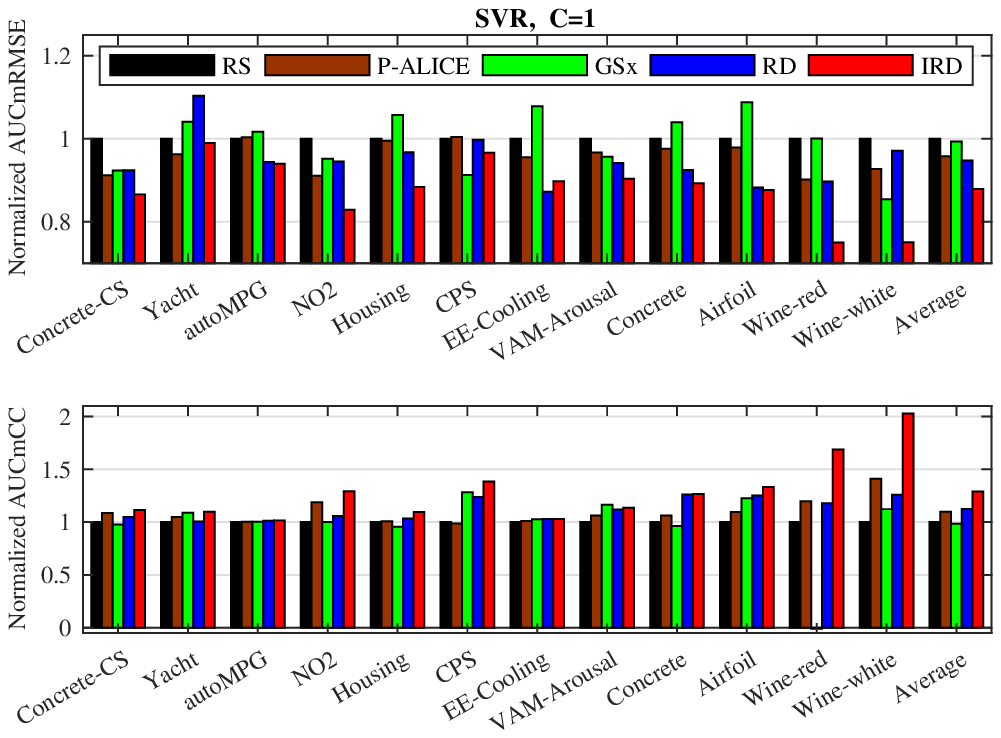}}
\caption{Normalized AUCs of the mean RMSEs and the mean CCs on the 12 datasets. (a) RR ($\lambda=0.5$); (b) LASSO ($\lambda=0.5$); and, (c) linear SVR ($C=1$).} \label{fig:bar}
\end{figure}

To see the forest for the trees, we also computed the area under the curves (AUCs) of the mean RMSE and the mean CC (AUC-mRMSE and AUC-mCC) and show the results in Fig.~\ref{fig:barRR}. Note that the AUCs on different datasets differed significantly on their magnitude, so it's challenging to show their raw values in a single plot. To accommodate this, for each dataset, we normalized the AUCs of the four ALR approaches w.r.t. that of RS, and hence the normalized AUC of RS was always one in Fig.~\ref{fig:barRR}. The last group in the subfigure shows the average normalized AUCs across the 12 datasets.

Fig.~\ref{fig:barRR} indicates that:
\begin{enumerate}
\item IRD achieved the best RMSE on 10 out of the 12 datasets, the second best on the remaining two, and hence the best average RMSE on all 12 datasets. It also achieved the highest CC on 10 datasets, the second/third highest on the remaining two, and hence the highest average CC on all 12 datasets.
\item On average RD slightly outperformed P-ALICE, both of which outperformed RS.
\item GSx had the worst RMSE on seven datasets, the second worst on another three, and hence the worst average RMSE on all 12 datasets. It also had the lowest CC on six datasets, and hence the lowest average CC on all 12 datasets.
\end{enumerate}
In summary, the rank of the performances of the five unsupervised sampling approaches, from the best to the worst, was IRD$>$RD$>$P-ALICE$>$RS$>$GSx.

The AUCs of the five unsupervised sampling approaches, from the three regression models, averaged across the 12 datasets, are shown in Table~\ref{tab:imp}. The reason why GSx did not work well when $M$ was small may be that the samples selected by GSx were mostly outliers. The negative effect of outliers may out-weighted the positive effect of increased diversity in regression. IRD performed the best because it simultaneously considers informativeness, representativeness and diversity, whereas RD only considers representativeness and diversity, and P-ALICE only considers informativeness.

\begin{table}[h]   \centering \setlength{\tabcolsep}{1mm}
  \caption{Percentage improvements of the AUCs of the mean RMSEs and the mean CCs.}   \label{tab:imp}
    \begin{tabular}{c|c|l|cccc}
    \toprule
    Regression & \multicolumn{2}{c|}{Performance}& \multicolumn{4}{c}{Percentage Improvement Relative to RS} \\ \cline{4-7}
    Model & \multicolumn{2}{c|}{Measure} & P-ALICE & GSx & RD & IRD \\    \midrule
    \multirow{4}[4]{*}{RR} & \multirow{2}[2]{*}{RMSE} & Mean & 2.58  & -2.57  & 4.15  & \textbf{8.63}  \\
      &   & std & 2.75  & 3.98  & \textbf{36.60}  & 34.84  \\
\cmidrule{2-7}      & \multirow{2}[2]{*}{CC} & Mean & 6.54  & -3.43  & 10.39  & \textbf{18.70}  \\
      &   & std & 12.74  & 29.47  & 35.03  & \textbf{42.97}  \\
    \midrule
    \multirow{4}[4]{*}{LASSO} & \multirow{2}[2]{*}{RMSE} & Mean & 4.22  & 0.84  & 7.58  & \textbf{10.81}  \\
      &   & std & 6.77  & 0.85  & \textbf{43.45}  & 39.84  \\
\cmidrule{2-7}      & \multirow{2}[2]{*}{CC} & Mean & 25.06  & \textbf{69.41}  & 25.67  & 60.63  \\
      &   & std & 6.39  & \textbf{31.05}  & 22.46  & 29.82  \\
    \midrule
    \multirow{4}[4]{*}{SVR} & \multirow{2}[2]{*}{RMSE} & Mean & 4.21  & 0.66  & 5.23  & \textbf{12.12}  \\
      &   & std & 6.62  & -0.19  & 33.99  & \textbf{38.69}  \\
\cmidrule{2-7}      & \multirow{2}[2]{*}{CC} & Mean & 9.71  & -1.65  & 12.46  & \textbf{28.99}  \\
      &   & std & 11.10  & 25.78  & 34.97  & \textbf{43.25}  \\
    \bottomrule
    \end{tabular}
\end{table}

It is also interesting to study the consistency of the sample selection approaches. Given similar average performances, an algorithm with a smaller variation is usually preferred in practice. Table~\ref{tab:imp} shows the improved standard deviations (std) of the AUCs of the RMSEs and the CCs from the 100 runs, averaged across the 12 datasets. Again, on average IRD had the largest improvement on the std, i.e., it was the most consistent ALR approach.

\begin{figure}[htbp]\centering
\subfigure[]{\label{fig:RatioRR}   \includegraphics[width=\linewidth,clip]{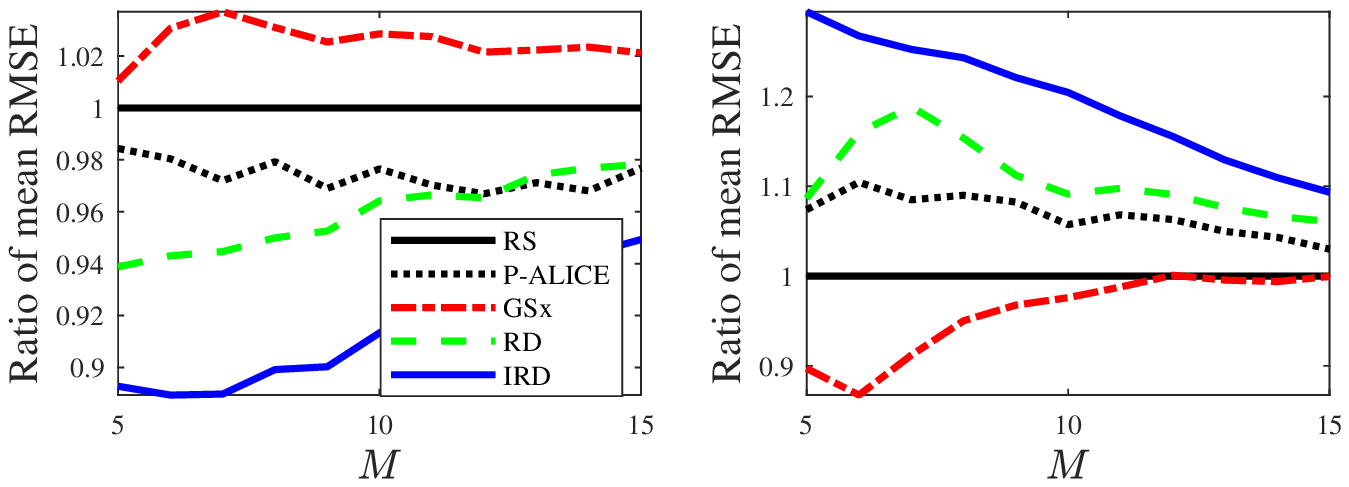}}
\subfigure[]{\label{fig:RatioLASSO}   \includegraphics[width=\linewidth,clip]{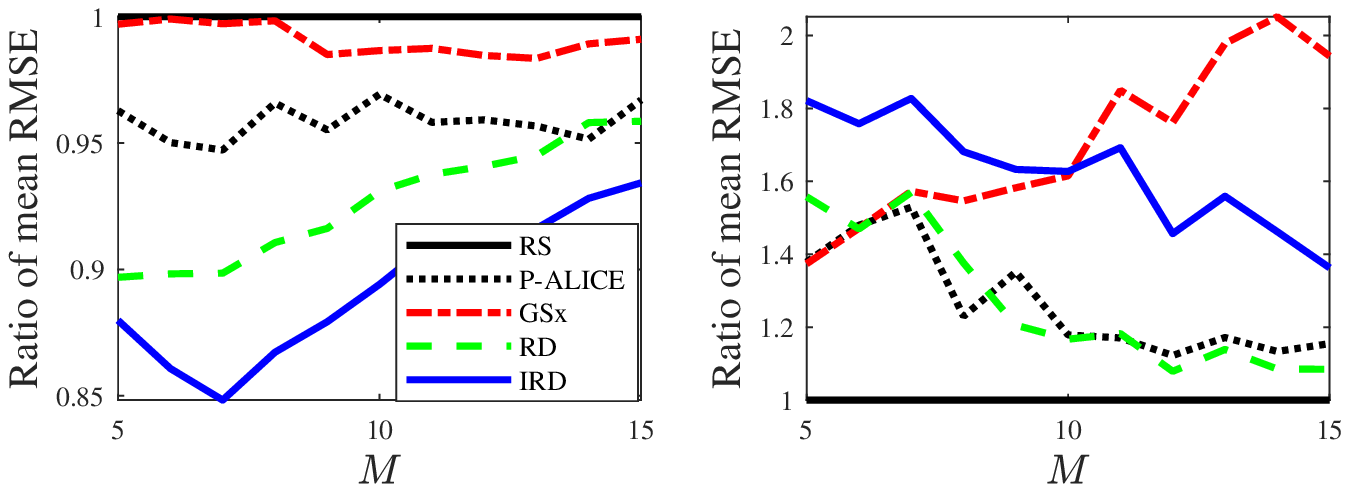}}
\subfigure[]{\label{fig:RatioSVR}   \includegraphics[width=\linewidth,clip]{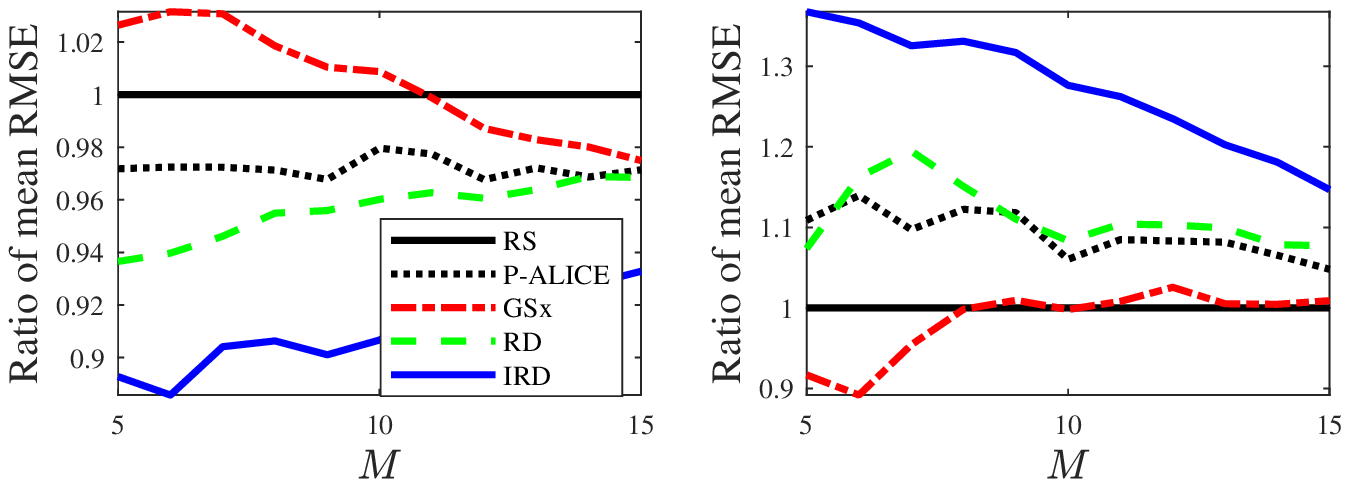}}
\caption{Ratios of the mean RMSEs and the mean CCs for different $M$, averaged across 12 datasets. (a) RR ($\lambda=0.5$); (b) LASSO ($\lambda=0.5$); and, (c) linear SVR ($C=1$).} \label{fig:Ratios}
\end{figure}

For each $M$ in each run on each dataset, we also computed the ratio of the corresponding RMSEs (CCs) of P-ALICE, GSx, RD and IRD to that of RS, and then the mean of the ratios for each $M$, across the 100 runs and the 12 datasets. The results are shown in Fig.~\ref{fig:Ratios}. The performance improvements of IRD over the other four approaches were particularly large when $M$ was small, thanks to the innovative way IRD uses to evaluate the informativeness and the diversity. As $M$ increased, the performance improvements of all four ALR approaches over RS decreased, which is intuitive, as when the number of labeled samples increases, the impact of the optimality of each sample decreases.

\subsection{Experimental Results Using LASSO and Linear SVR} \label{sect:LASSOSVR}

We also repeated the above experiments and analyses when LASSO and linear SVR were used as the linear regression model. The results are shown in Figs.~\ref{fig:barLASSO} and \ref{fig:barSVR}. They demonstrate similar patterns as those in Fig.~\ref{fig:barRR}, e.g., IRD always achieved the best average performance, and RD outperformed P-ALICE, RS and GSx. Moreover, the performance improvements of the four ALR approaches, particularly IRD, over RS, were generally more obvious than those for RR.

To quantify the performance improvements of the four unsupervised ALR approaches over RS, we computed the percentage improvements on the AUCs of the RMSEs and the CCs, as shown in Table~\ref{tab:imp}. It confirmed that on average IRD outperformed all other four approaches, regardless of which linear regression model was trained, and which performance measure was used.

\subsection{Statistical Analysis}

To determine if the performance differences between IRD and the other four approaches were statistically significant, we also performed non-parametric multiple comparison tests on them using Dunn's procedure \cite{Dunn1961}, with a $p$-value correction using the False Discovery Rate method \cite{Benjamini1995}. The results are shown in Table~\ref{tab:stat}, where the statistically significant ones are marked in bold.

\begin{table}[h]  \centering \setlength{\tabcolsep}{2mm}
  \caption{$p$-values of non-parametric multiple comparisons on the AUCs of RMSEs and CCs ($\alpha=0.05$; reject $H_0$ if $p<\alpha/2$). The statistically significant ones are marked in bold.}  \label{tab:stat}
    \begin{tabular}{c|c|cccc}
    \toprule
    Regression & Performance & \multicolumn{4}{c}{IRD versus} \\ \cline{3-6}
    Model & Measure &  RS & P-ALICE & GSx & RD \\
    \midrule
    \multirow{2}[2]{*}{RR} & RMSE & \textbf{.0000} & \textbf{.0003} & \textbf{.0000} & .0284  \\
      & CC & \textbf{.0000} & \textbf{.0000} & \textbf{.0000} & \textbf{.0005} \\
    \midrule
    \multirow{2}[2]{*}{LASSO} & RMSE & \textbf{.0000} & \textbf{.0004} & \textbf{.0000} & .0596  \\
      & CC & \textbf{.0000} & \textbf{.0000} & \textbf{.0000} & \textbf{.0000} \\
    \midrule
    \multirow{2}[2]{*}{SVR} & RMSE & \textbf{.0000} & \textbf{.0000} & \textbf{.0000} & \textbf{.0018} \\
      & CC & \textbf{.0000} & \textbf{.0000} & \textbf{.0000} & \textbf{.0000} \\
    \bottomrule
    \end{tabular}%
\end{table}%

No matter which linear regression model was used, the RMSE and CC improvements of IRD over RS, P-ALICE and GSx were always statistically significant. The CC improvements of IRD over RD were also always statistically significant, and the RMSE improvement was statistically significant when linear SVR was used.

\subsection{Visualization of the Selected Samples}

\begin{figure}[h]\centering
\includegraphics[width=\linewidth,clip]{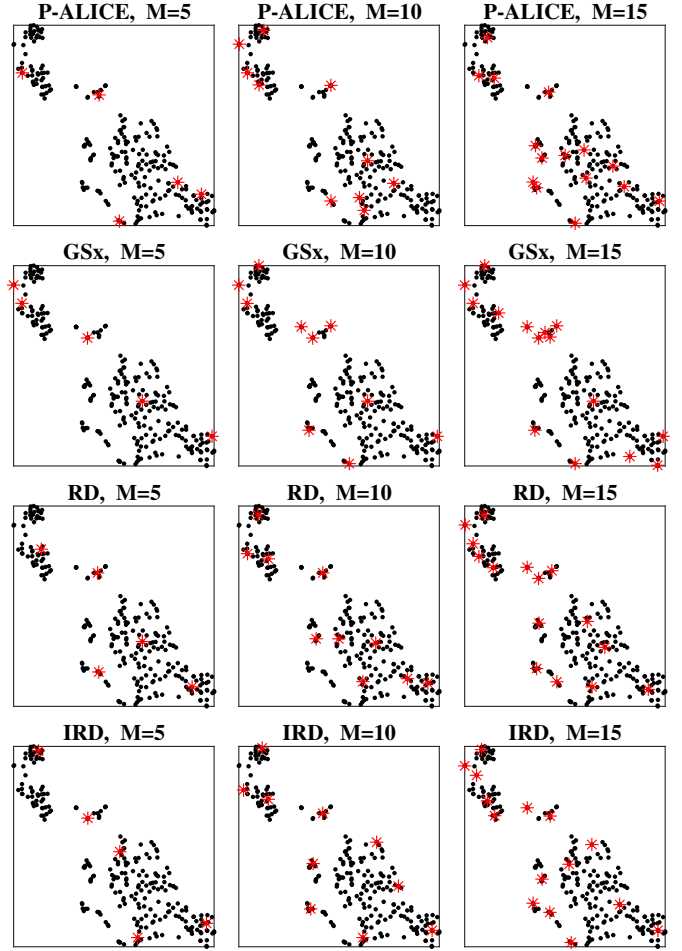}
\caption{t-SNE visualization of the selected samples (red asterisks) from different ALR approaches on the Housing dataset.} \label{fig:IRD_tsne_Housing}
\end{figure}

To visualize the differences among the selected samples by different ALR approaches, we picked a typical dataset (Housing), used t-SNE \cite{Maaten2008} to map the samples to 2D, and then show the selected samples from the four ALR approaches for three different $M$ in Fig.~\ref{fig:IRD_tsne_Housing}. Note that there is also a weight associated with each sample selected by P-ALICE, which is ignored in the plots.

Fig.~\ref{fig:IRD_tsne_Housing} shows that GSx tended to select the boundary samples, which may be outliers. Additionally, the distribution of the selected samples was inconsistent with those in the pool. As a result, its performance was on average the worst among the four. The samples selected by P-ALICE and RD were more uniformly distributed in the pool than those by GSx. The samples selected by IRD tended to be near the boundary of the pool, but not exactly at the boundary, which were less likely to be outliers. Additionally, the distribution of its selected samples was consistent with those of the pool. These may contribute to its outstanding performance.

\subsection{Influence of $c_{\max}$ on IRD}

Algorithms~1-3 have an important parameter, $c_{\max}$, the maximum number of iterations. IRD is equivalent to RD when $c_{\max}=0$. This subsection studies whether by setting $c_{\max}>0$, IRD can indeed improve over RD.

\begin{figure}[h]\centering
\subfigure[]{\label{fig:cRR}   \includegraphics[width=\linewidth,clip]{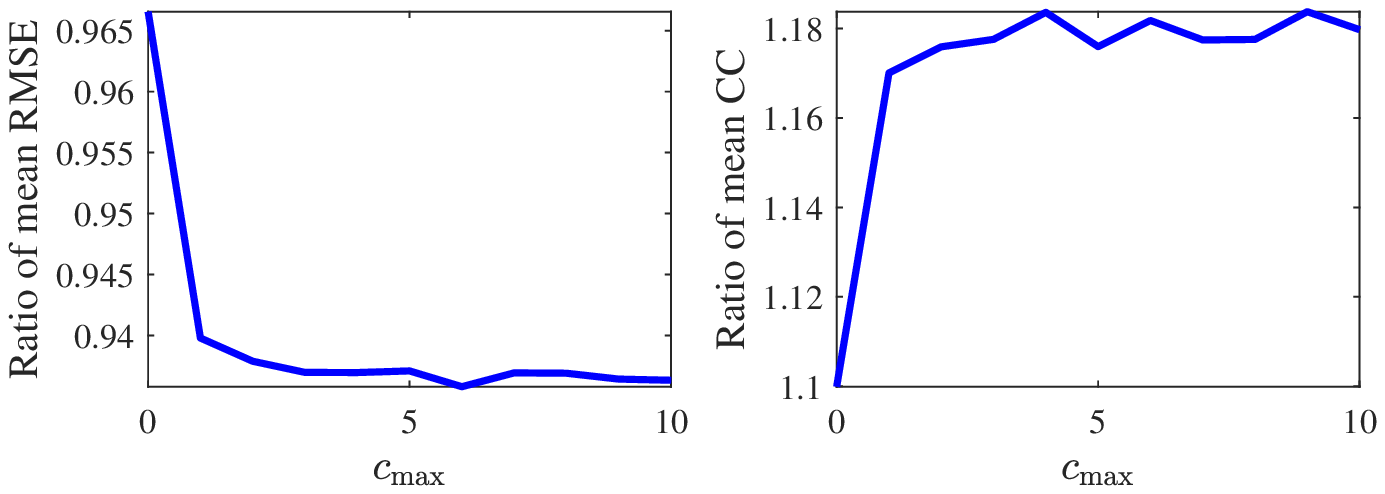}}
\subfigure[]{\label{fig:cLASSO}   \includegraphics[width=\linewidth,clip]{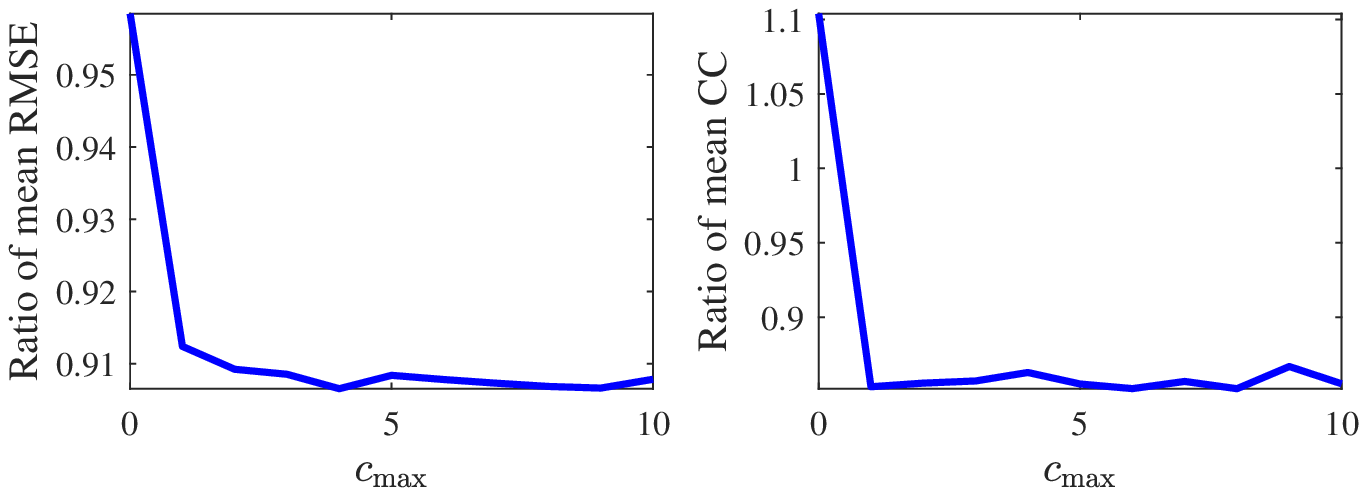}}
\subfigure[]{\label{fig:cSVR}   \includegraphics[width=\linewidth,clip]{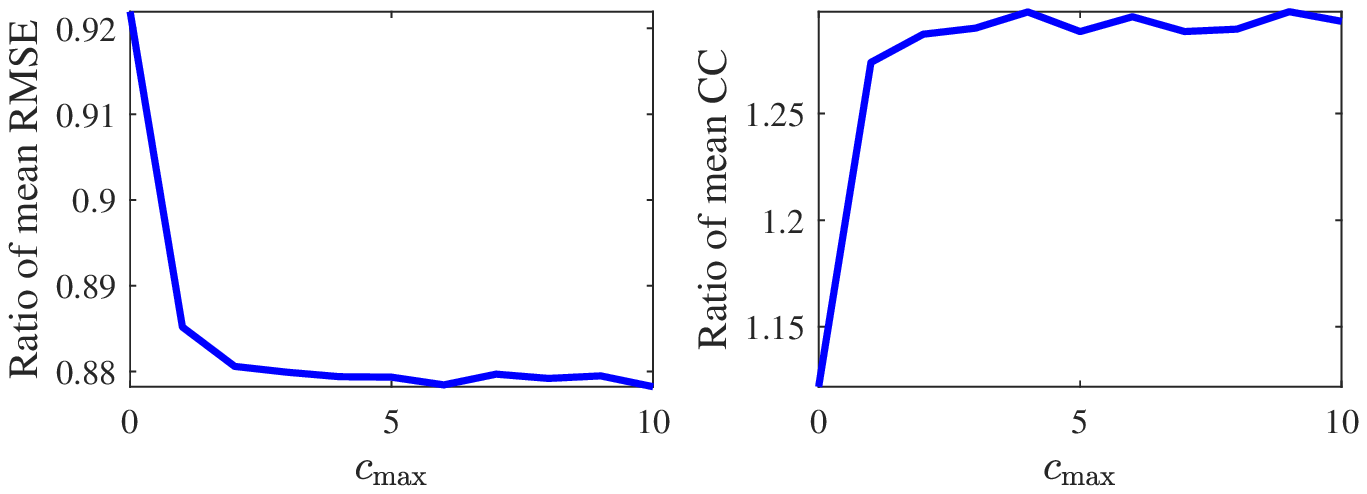}}
\caption{Ratios of AUCs of the mean RMSEs and the mean CCs for different $c_{\max}$, averaged across 12 datasets. (a) RR ($\lambda = 0.5$); (b) LASSO ($\lambda = 0.5$); and, (c) linear SVR ($C=1$).} \label{Fig:Cmax}
\end{figure}

Fig.~\ref{Fig:Cmax} shows the normalized AUCs w.r.t. RS, averaged across 100 runs and 12 datasets, using three different linear regression models and $c_{\max}\in[0,10]$. The performance of IRD improved quickly as $c_{\max}$ increased and always reached the optimum before $c_{\max}=5$, which means the iterative approach in IRD was both effective and efficient.

\subsection{Influence of the Regularization Coefficients}

\begin{figure}[h]\centering
\subfigure[]{\label{fig:lambdaRR}   \includegraphics[width=\linewidth,clip]{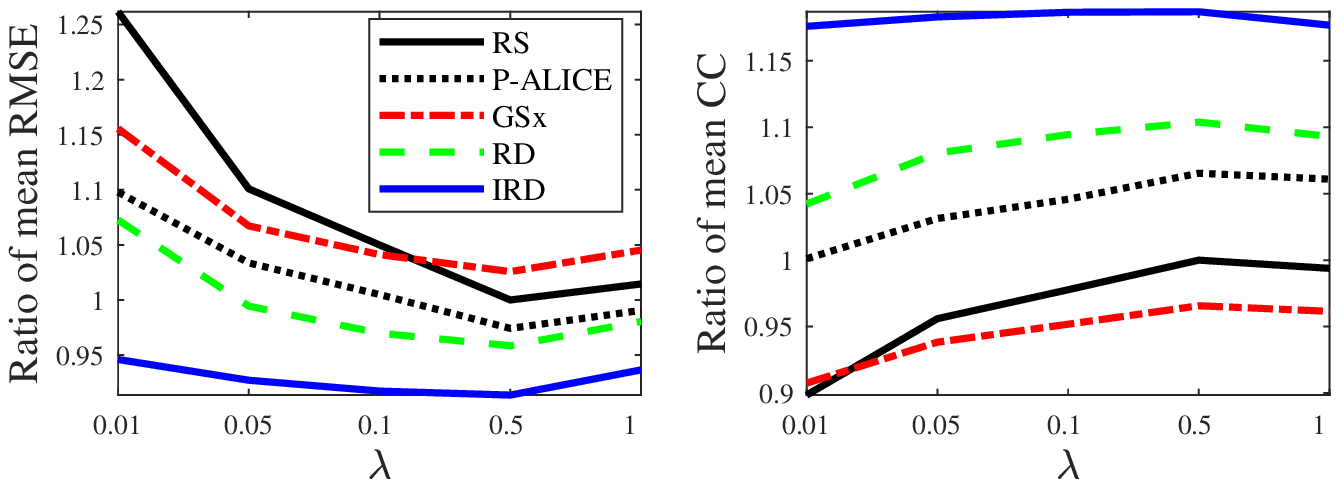}}
\subfigure[]{\label{fig:lambdaLASSO}   \includegraphics[width=\linewidth,clip]{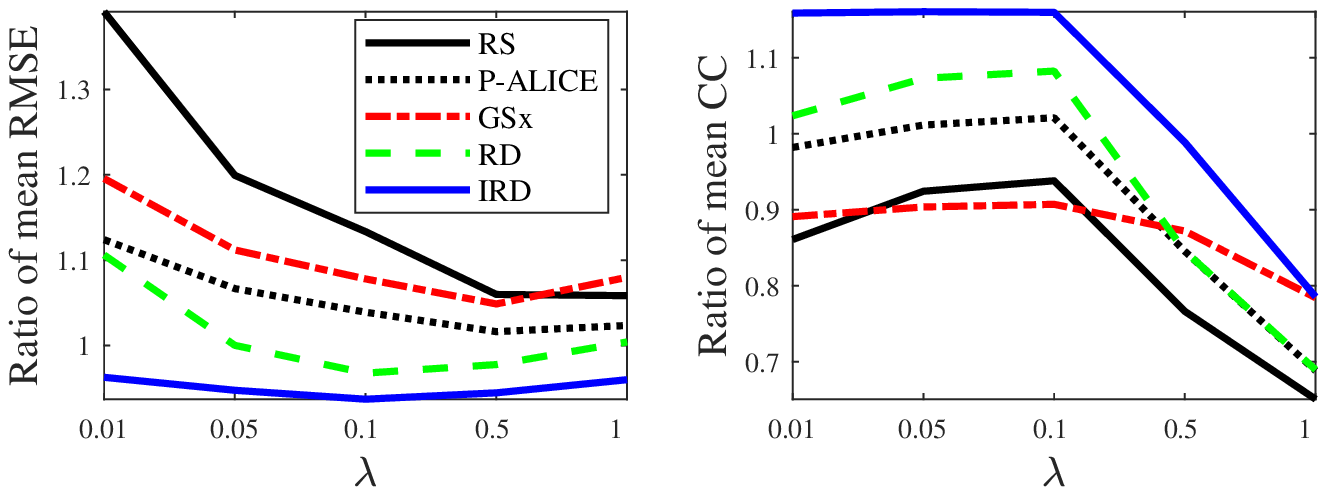}}
\subfigure[]{\label{fig:lambdaSVR}   \includegraphics[width=\linewidth,clip]{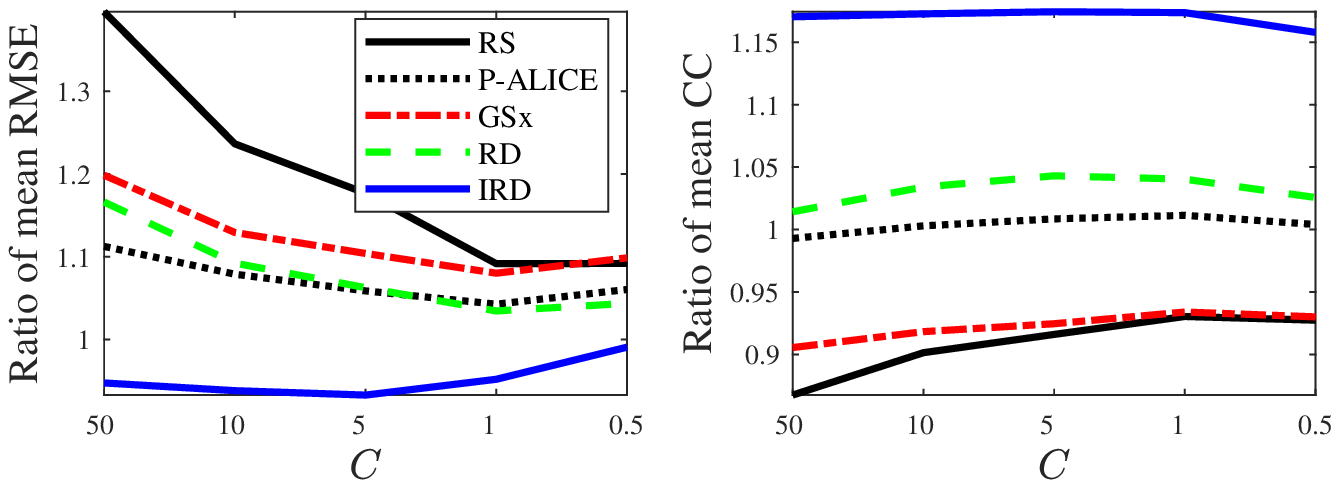}}
\caption{Ratios of the AUCs of the mean RMSEs and the mean CCs, averaged across 12 datasets, for (a) different $\lambda$ in RR; (b) different $\lambda$ in LASSO; and, (c) different $C$ in linear SVR.} \label{fig:lambda}
\end{figure}

To study whether the performances of the five unsupervised sampling approaches were sensitive to the regularization coefficients of the three linear regression models, we repeated the experiments for $\lambda\in\{0.01,0.05,0.1,0.5,1\}$. Note that linear SVR has an equivalent L2 regularization coefficient $\lambda = \frac{1}{2C}$, so we set $C\in\{50, 10, 5, 1, 0.5\}$. For each dataset, we normalized the AUCs of each sampling approach and each regression model w.r.t. that of RS (using ridge parameter $\lambda=0.5$). Fig.~\ref{fig:lambda} shows the normalized AUCs of each approach and each regression model, averaged across 100 runs and 12 datasets.

The performances of the five unsupervised sampling approaches first improved as $\lambda$ increased, and then decreased. However, regardless of the value of $\lambda$ ($C$), generally IRD always performed the best, and RD the second best. The performance improvements of IRD over the other four approaches were larger for smaller $\lambda$. Moreover, IRD was not very sensitive to $\lambda$, which is desirable in real-world applications.

\subsection{Individual Contributions of Informativeness, Representativeness, and Diversity}

This subsection studies the individual effects of informativeness, representativeness and diversity, by comparing the following variants of IRD:
\begin{enumerate}
\item IRD ($c_{\max}=5$), which is our proposed approach, introduced in Section~\ref{sect:IRD}.
\item ID, which considers only the denominator of (\ref{eq:IRD}) when $M=d+1$, and only $D(\mathbf{x}_n)$ in (\ref{eq:IRDr}) when $M>d+1$, i.e., only the informativeness and the diversity.
\item RD, which is IRD with $c_{\max}=0$ and uses RD in the initialization. This approach is equivalent to RD in \cite{drwuSAL2019}, i.e., it considers only the representativeness and the diversity.
\end{enumerate}
Each approach was run on each of the 12 datasets for 100 times for $M\in[5, 15]$. RR ($\lambda = 0.5$), LASSO ($\lambda = 0.5$) and Linear SVR ($C=1$) were used as the regression models for each $M$.

\begin{figure}[h]\centering
\subfigure[]{\label{fig:mRR}   \includegraphics[width=\linewidth,clip]{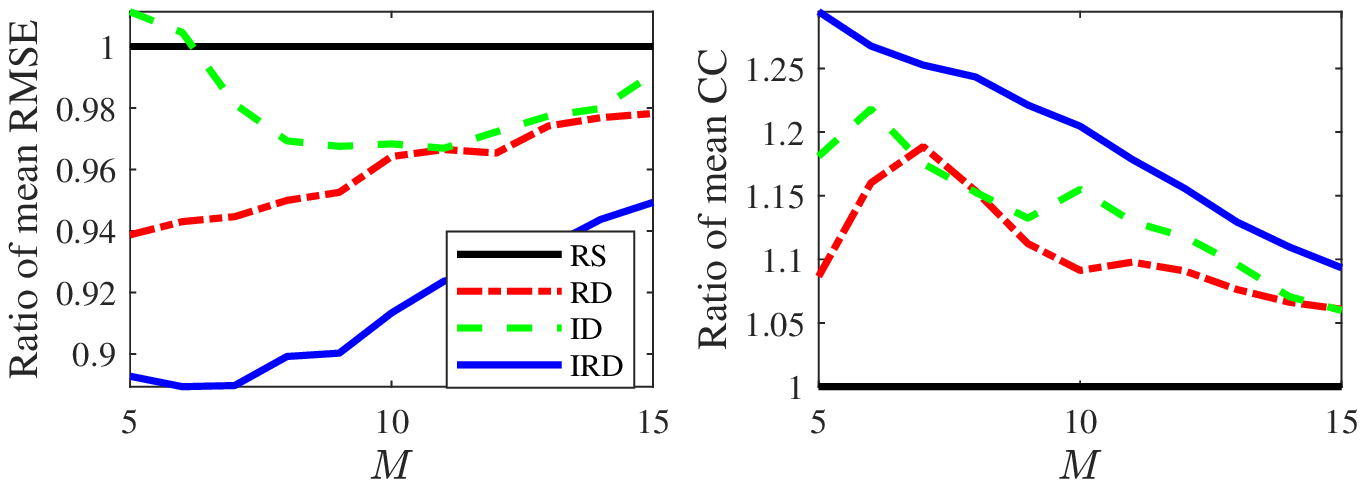}}
\subfigure[]{\label{fig:mLASSO}   \includegraphics[width=\linewidth,clip]{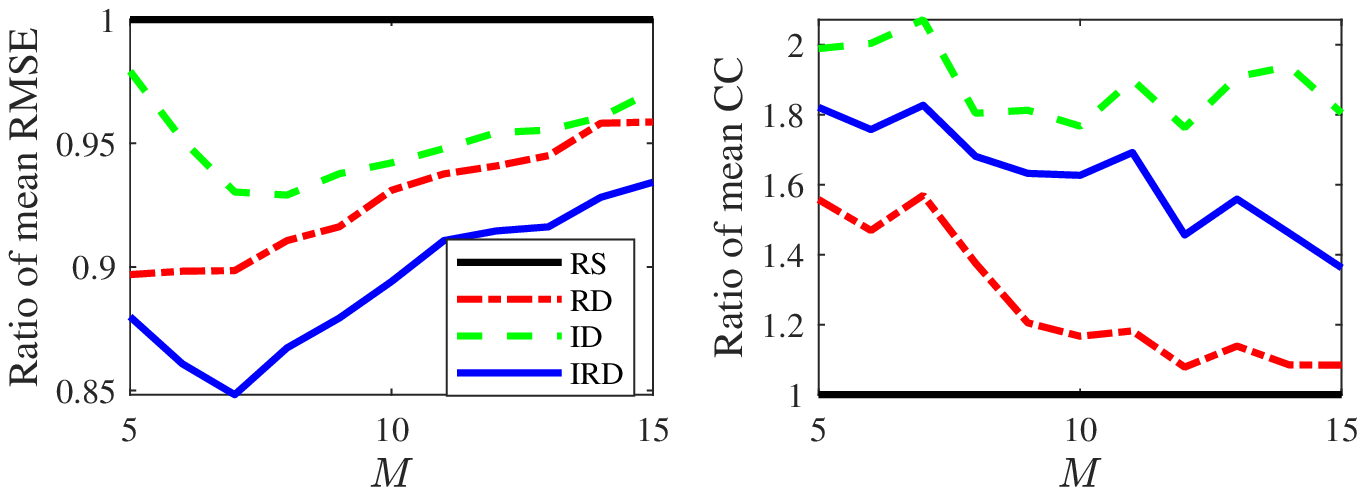}}
\subfigure[]{\label{fig:mSVR}   \includegraphics[width=\linewidth,clip]{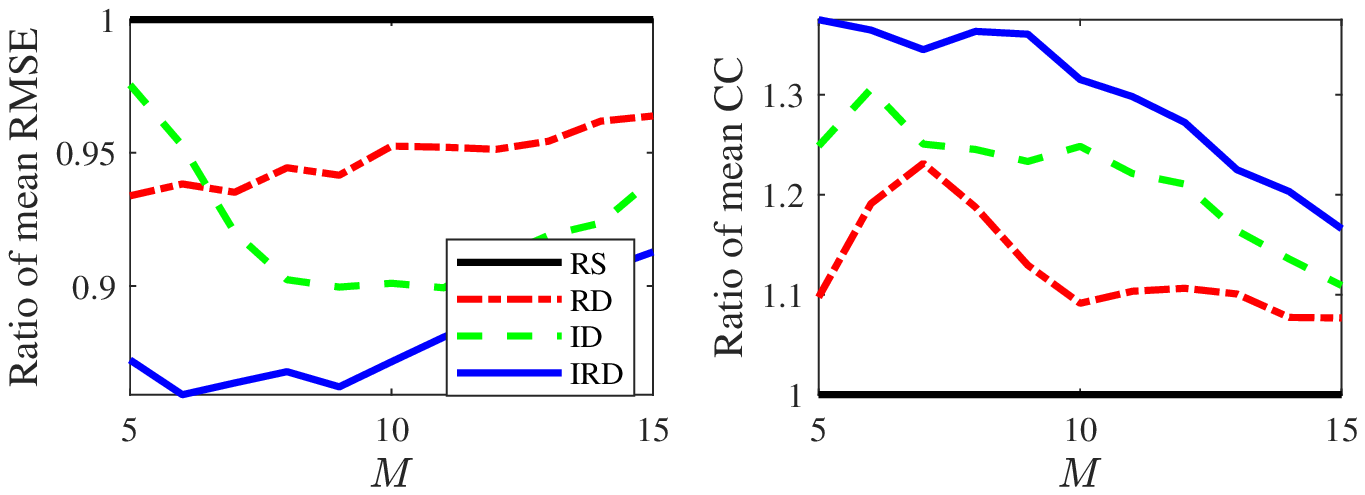}}
\caption{Ratios of the mean RMSEs and the mean CCs w.r.t. different $M$, averaged across 12 datasets. (a) RR ($\lambda = 0.5$); (b) LASSO ($\lambda = 0.5$); and, (c) linear SVR ($C=1$).} \label{fig:M}
\end{figure}

Fig.~\ref{fig:M} shows the ratios of the mean RMSEs and the mean CCs w.r.t. those of RS for each $M$, averaged across 100 runs and 12 datasets. The results from the three regression models were similar. Generally, all three ALR approaches outperformed RS. However, IRD still performed the best, suggesting that it is critical to consider informativeness, representativeness and diversity simultaneously.

\section{Conclusions} \label{sect:conclusions}

AL aims to select the optimal samples to label, so that a machine learning model built from them can achieve the best possible performance. Thus, it is very useful in real-world applications, where data labeling is time-consuming and/or expensive. Most existing AL approaches are supervised: they train an initial model from a small amount of labeled data, query new data based on the model, and then update the model. This paper considered completely unsupervised pool-based AL problems for linear regression, i.e., how to optimally select the very first few samples to label, without knowing any label information at all. We proposed a novel AL approach that considers simultaneously three essential criteria in AL: informativeness, representativeness, and diversity. Extensive experiments on 14 datasets from various application domains, using three different linear regression models (RR, LASSO, and linear SVR), demonstrated that our proposed approach significantly outperformed three state-of-the-art unsupervised pool-based AL approaches for linear regression.


\end{document}